\def\year{2022}\relax
\documentclass[letterpaper]{article} %
\usepackage{aaai22}  %
\usepackage{times}  %
\usepackage{helvet}  %
\usepackage{courier}  %
\usepackage[hyphens]{url}  %
\usepackage{graphicx} %
\usepackage{natbib}  %
\usepackage{caption} %
\usepackage{amsmath}
\usepackage{amssymb}
\usepackage{epsfig}
\usepackage{pifont}
\usepackage{subfigure}
\usepackage{booktabs}
\usepackage{hyperref}
\hypersetup{
    colorlinks = true,
    citecolor = black
}
\usepackage{hhline}
\usepackage[table]{xcolor}
\usepackage{color, colortbl}
\usepackage{multirow}
\usepackage{algorithm}
\usepackage{algorithmic}
\DeclareCaptionStyle{ruled}{labelfont=normalfont,labelsep=colon,strut=off} %
\usepackage{algorithm}
\usepackage{algorithmic}
\usepackage[switch]{lineno}

\usepackage{newfloat}
\usepackage{listings}
\floatstyle{ruled}
\newfloat{listing}{tb}{lst}{}
\floatname{listing}{Listing}
\usepackage{listings}
\title{Contrastive Instruction-Trajectory Learning for Vision-Language Navigation}
\author{
Xiwen Liang\textsuperscript{\rm 1},
Fengda Zhu\textsuperscript{\rm 2},
Yi Zhu\textsuperscript{\rm 3},
Bingqian Lin\textsuperscript{\rm 1},
Bing Wang\textsuperscript{\rm 4},
Xiaodan Liang\textsuperscript{\rm 1}\thanks{Corresponding author.}
}
\affiliations{
\textsuperscript{\rm 1}Shenzhen Campus of Sun Yat-sen University, Shenzhen
\textsuperscript{\rm 2}Monash University
\textsuperscript{\rm 3}Huawei Noah's Ark Lab
\textsuperscript{\rm 4}Alibaba group\\
liangxw29@mail2.sysu.edu.cn, fengda.zhu@monash.edu, zhu.yee@outlook.com, linbq6@mail2.sysu.edu.cn, fengquan.wb@alibaba-inc.com, liangxd9@mail.sysu.edu.cn
}

\usepackage{bibentry}

\begin{document}

\maketitle

\begin{abstract}
The vision-language navigation (VLN) task requires an agent to reach a target with the guidance of natural language instruction. 
Previous works learn to navigate step-by-step following an instruction. 
However, these works may fail to discriminate the similarities and discrepancies across instruction-trajectory pairs and ignore the temporal continuity of sub-instructions.
These problems hinder agents from learning distinctive vision-and-language representations, harming the robustness and generalizability of the navigation policy. 
In this paper, we propose a Contrastive Instruction-Trajectory Learning (CITL) framework that explores invariance across similar data samples and variance across different ones to learn distinctive representations for robust navigation. 
Specifically, we propose: 
(1) a coarse-grained contrastive learning objective to enhance vision-and-language representations by contrasting semantics of full trajectory observations and instructions, respectively; 
(2) a fine-grained contrastive learning objective to perceive instructions by leveraging the temporal information of the sub-instructions; 
(3) a pairwise sample-reweighting mechanism for contrastive learning to mine hard samples and hence mitigate the influence of data sampling bias in contrastive learning.
Our CITL can be easily integrated with VLN backbones to form a new learning paradigm and achieve better generalizability in unseen environments. 
Extensive experiments show that the model with CITL surpasses the previous state-of-the-art methods on R2R, R4R, and RxR.
\end{abstract}

\section{Introduction}

\begin{figure*}
    \centering
    \includegraphics[width=1.0\linewidth]{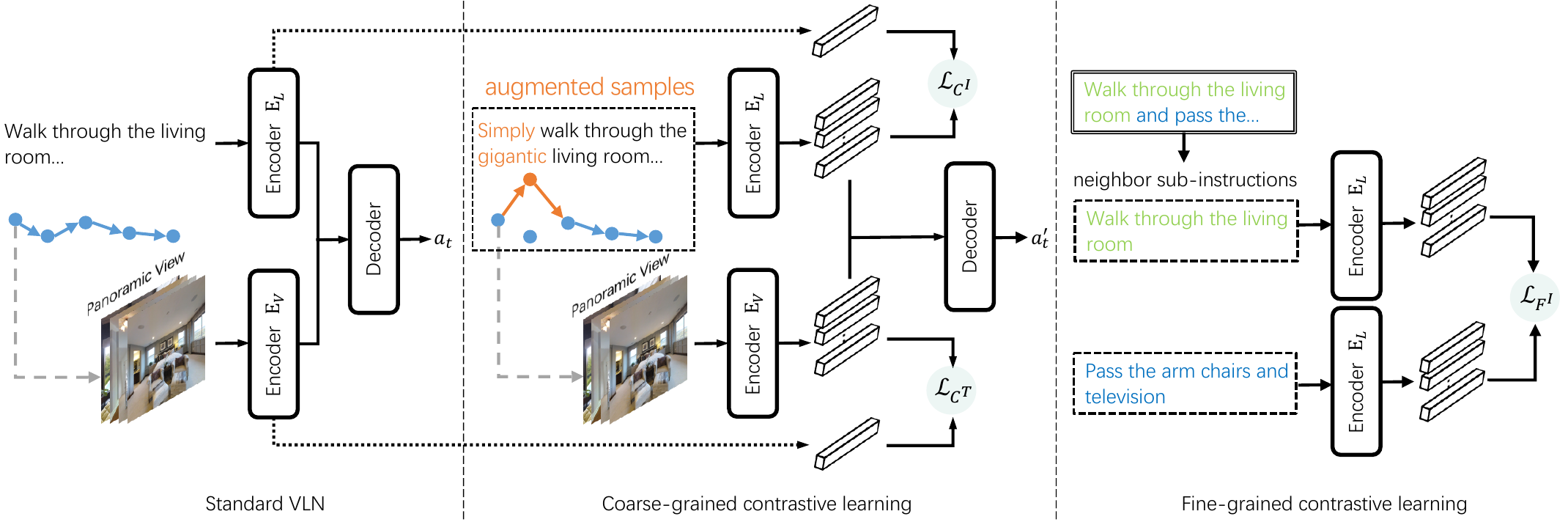}
    \caption{We propose contrastive instruction-trajectory learning for VLN (middle and right). Most previous works use a follower model to produce actions in VLN (left), neglecting different transformations of the instruction-trajectory pairs. As a result, representations may be variant with similar instruction-trajectory pairs. By contrast, our CITL learns representations similar to the input instruction-trajectory pair with different transformations (middle) and retains rich semantic information. Our CITL also leverages the temporal continuity of the sub-instructions (right).}
    \vspace{-1 em}
    \label{fig:compare}
\end{figure*}

Vision-Language Navigation (VLN) task~\cite{Anderson2018Vision} requires an agent to navigate following a natural language instruction. 
This task is closely connected to many real-world applications, such as household robots and rescue robots. 
The VLN task is challenging since it requires an agent to acquire diverse skills, such as vision-language alignment, sequential vision perception and long-term decision making. 

Early method~\cite{Anderson2018Vision} is developed upon an encoder-decoder framework~\cite{Sutskever2014Sequence}.
Later methods~\cite{Fried2018Speaker,Wang2019Reinforced, Zhu2020Vision, Ke2019Tactical, ma2019selfmonitoring} improve the agent with vision-language attention layers and auxiliary tasks.
Coupling with BERT-like methods~\cite{Jacob2019BERT, Lu2019ViLBERT, li2020oscar}, the navigation agent obtains better generalization ability~\cite{majumdar2020improving, hong2020recurrent}. However, these VLN methods only use the context within an instruction-trajectory pair while ignoring the knowledge across the pairs. 
For instance, they only recognize the correct actions that follow the instruction while ignoring the actions that do not follow the instruction. 
The differences between the correct actions and the wrong actions contain extra knowledge for navigation. 
On the other hand, previous methods do not explicitly exploit the temporal continuity inside an instruction, which may fail if the agent focuses on a wrong sub-instruction. Thus, learning a fine-grained sub-instruction representation by leveraging the temporal continuity of sub-instructions could improve the robustness of navigation. 

Recently, self-supervised contrastive learning shows superior capacity in improving the instance discrimination and generalization of vision models~\cite{Chen2020simple,He2020Momentum,xie2020propagate,li2020unimo,sun2019learning}. 
Inspired by the success of contrastive learning, we propose our Contrastive Instruction-Trajectory Learning (CITL) framework to explore fine/coarse-grained knowledge of the instruction-trajectory pairs. 
Our CITL consists of two coarse-grained trajectory-instruction contrastive objectives and a 
fine-grained sub-instruction contrastive objective to learn from cross-instance trajectory-instruction pairs and sub-instructions. 
Firstly, we propose \textbf{coarse-grained contrastive learning} to learn distinctive long-horizon representations for trajectories and instructions respectively. 
The idea of coarse-grained contrastive learning is computing inter-intra cross-instance contrast: enforcing embedding to be similar for positive trajectory-instruction pairs and dissimilar for intra-negative and inter-negative ones.
To obtain positive samples, we propose data augmentation methods for instructions and trajectories respectively. Intra-negative samples are generated through changing the temporal information of the instruction and selecting longer sub-optimal trajectories which deviate from the anchor one severely. In contrast, inter-negative samples are different trajectory-instruction pairs.
In this way, the semantics of full trajectory observations and instructions can be captured for better shaping representations with less variance under diverse data transformations.
Secondly, we propose \textbf{fine-grained contrastive learning} to learn fine-grained representations by focusing on the temporal information of sub-instructions. 
We generate sub-instructions as in~\cite{hong2020sub}, and train the agent to learn embedding distances of these sub-instructions by contrastive learning. 
Specifically, neighbor sub-instructions are positive samples, while non-neighbor sub-instructions are intra-negative samples and different sub-instructions from other instructions are inter-negative samples.
These learning objectives help the agent leverage richer knowledge to learn better embedding for instructions and trajectories,  
and therefore, obtain a more robust navigation policy and better generalizability. 
Fig.~\ref{fig:compare} shows an overview of our CITL framework. 

We also overcome several challenges in adopting contrastive learning in VLN by introducing \textbf{pairwise sample-reweighting mechanism}. 
Firstly, a large scale of easy samples dominates the gradient, causing the performance to plateau quickly. Some false-negative samples may exist and introduce noise. To avoid these problems, we introduce pair mining to mine hard samples and remove false-negative ones online, making the model focuses on hard samples during training.
Secondly, the generated positive trajectories may be close to or heavily deviate from the anchor one. Previous multi-pair contrastive learning  methods~\cite{xie2020pointcontrast,cai2020joint} adopt InfoNCE loss~\cite{oord2018representation}, which fails to explicitly penalize samples differently. 
Therefore we introduce the circle loss~\cite{Sun2020Circle} to penalize different positive and negative samples.

Our experiments demonstrate that our CITL framework can be easily combined with different VLN models and significantly improves their navigation performance (2\%-4\% in terms of SPL in R2R and R4R). 
Our ablation studies show that CITL helps the model learn more distinct knowledge with different data transformations since coarse/fine-grained contrastive objectives introduce cross-instance long-horizon information and intra-instance fine-grained information.

\begin{figure*}
    \centering
    \includegraphics[width=0.95\textwidth]{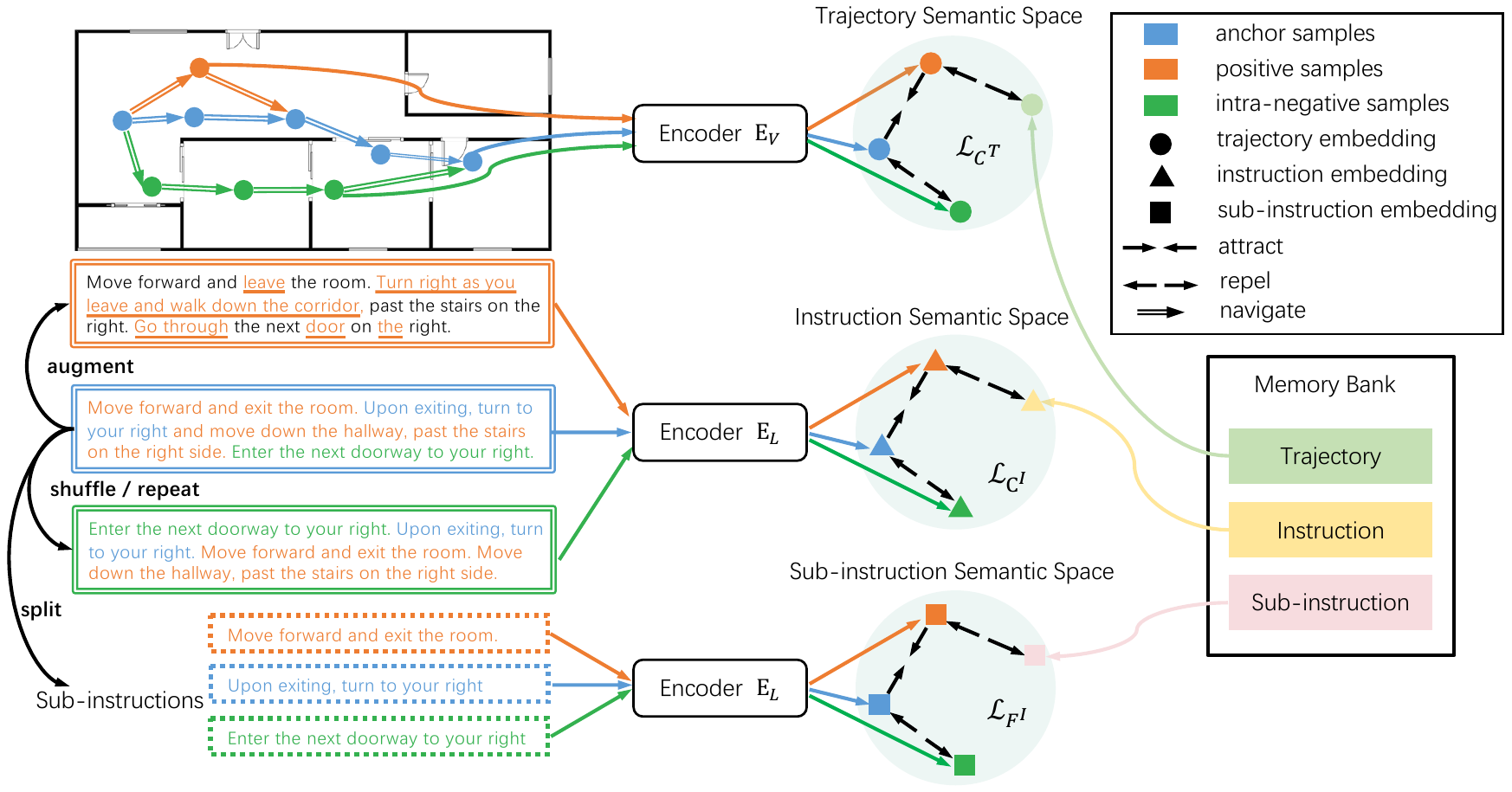}
    \caption{Overview of our coarse-fine contrastive learning-based vision-and-language navigation framework. Both trajectories and instructions are utilized for contrastive learning. Here trajectories are augmented by generating sub-optimal ones, and intra-negative trajectories deviate far away from the anchor one. Positive instructions are produced by back translation, inserting meaningful words and substituting with synonyms. Positive sub-instructions are neighboring to the anchor one. $\mathcal{L}_{C^T}$ is the coarse contrastive loss for trajectories, $\mathcal{L}_{C^I}$ represents the coarse contrastive loss for instructions, and $\mathcal{L}_{F^I}$ indicates the fine-grained contrastive loss for sub-instructions.}
    \label{fig:model}
    \vspace{-1 em}
\end{figure*}

\section{Related Work}

\noindent\textbf{Vision-and-Language Navigation} Learning navigation with vision-language clues has attracted a lot of attention of researchers. 
Room-to-Room (R2R)~\cite{Anderson2018Vision} and Touchdown~\cite{Chen2019TOUCHDOWNR} datasets introduce natural language and photo-realistic environment for navigation.
Following this, dialog-based navigation, such as VNLA~\cite{Nguyen2019Vision}, HANNA~\cite{nguyen2019hanna} and CVDN~\cite{Jesse2019Vision}, is proposed for further research. REVERIE~\cite{Qi2020REVERIE} introduces the task of localizing remote objects. 
A number of methods have been proposed to solve VLN. Speaker-Follower~\cite{Fried2018Speaker} introduces a speaker model and a panoramic representation to expand the limited data. Similarly, EnvDrop~\cite{Tan2019Learning} proposes a back-translation method to learn on augmented data. 
In~\cite{Ke2019Tactical}, an asynchronous search combined with global and local information is adopted to decide whether the agent should backtrack. To align the visual observation and the partial instruction better, a visual-textual co-grounding module is proposed in~\cite{ma2019selfmonitoring, Wang2019Reinforced}. Progress monitor and other auxiliary losses are proposed in~\cite{ma2019selfmonitoring,Ma2019TheRegretful, Zhu2020Vision, Qi2020Object, Wang2020Soft}. RelGraph~\cite{Hong2020Language} develops a language and visual relationship graph to model inter/intra-modality relationships. PRESS~\cite{Li2019RobustNW} applies the pre-trained BERT~\cite{Jacob2019BERT} to process instructions. RecBERT~\cite{hong2020recurrent} further implements a recurrent function based on ViLBERT. However, current VLN methods only focus on individual instruction-trajectory pairs and ignore the invariance of different data transformations. As a result, representations may be variant with similar instruction-trajectory pairs. 

\noindent\textbf{Contrastive Learning} Contrastive loss~\cite{Hadsell2006Dimensionality} is adopted to encourage representations to be close for similar samples and distant for dissimilar samples. Recently, state-of-the-art methods on unsupervised representation learning~\cite{Wu2018Unsupervised,He2020Momentum,Grill2020Bootstrap,Misra2020Self,caron2020unsupervised,Chen2020simple,chen2020exploring} are based on contrastive learning. Most methods adopt different transformations of an image as similar samples as in~\cite{Dosovitskiy2014Advances}.
Similar to contrastive loss, mutual information (MI) is maximized in~\cite{oord2018representation,Henaff2020DataEfficient,hjelm2019learning,Bachman2019Learning} to learn representations. In~\cite{hjelm2019learning,Bachman2019Learning,Henaff2020DataEfficient}, MI is maximized between global and local features from the encoder. \cite{Chaitanya2020Contrastive} integrates knowledge of medical imaging to define positive samples and focuses on distinguishing different areas in an image.
Memory bank~\cite{Wu2018Unsupervised} and momentum contrast~\cite{He2020Momentum,Misra2020Self} are proposed to use more negative pairs per batch. No work has attempted to life VLN models with the merits of contrastive learning. The success of contrastive learning motivates us to rethink the training paradigm of VLN and design contrastive learning objectives for VLN.

\noindent\textbf{Embedding Losses}  Contrastive loss~\cite{Hadsell2006Dimensionality} is a classic pair-based method in embedding learning. Triplet margin loss~\cite{Weinberger2005Distance} is proposed to capture variance in inter-class dissimilarities. Following these works, the margin of angular loss~\cite{Wang2017Deep} is based on angles of triplet vectors.
Lifted structure loss~\cite{Song2016Deep} applies LogSumExp, a smooth approximation of the maximum function, to all negative pairs. Softmax function is applied to each positive pair relative to all negative pairs in N-Pairs loss~\cite{Sohn2016Improved,oord2018representation,Chen2020simple}.
Similarities among each embedding and its neighbors are weighted explicitly or implicitly in~\cite{Wang2019Multi,Yu2019Deep,Sun2020Circle}.  
Unlike all previous work, our CITL is the first to adopt contrastive learning to learn distinct representations in VLN. Our proposed CITL differs from existing contrastive learning methods in several ways. Firstly, most previous single-modal contrastive learning approaches focus on image-level or pixel-level~\cite{xie2020propagate} comparison, and cross-modal contrastive learning methods mainly handle image-text pairs~\cite{li2020unimo} and video-text pairs~\cite{sun2019learning}, while we focus on trajectory-instruction pairs. Secondly, we introduce a pairwise sample-reweighting mechanism to learn trajectory-instruction representations effectively.

\section{Preliminaries}
\subsection{Vision-Language Navigation}
Given a natural language instruction $I$ with a sequence of words, at each time step $t$, the agent observes a panoramic view, which is divided into 36 single-view images $V_t=\{v_i\}_{i=1}^{36}$ for the agent to learn. 
The agent has $N_t$ navigable viewpoint as candidates, whose views from the current point are denoted as $C_t=\{c_i\}_{i=1}^{N_t}$. 
The agent predicts an action $a_t$ by selecting a viewpoint from $C_t$ to navigate each timestep.

A language encoder $\mathrm{E}_{L}$ and a vision encoder $\mathrm{E}_{V}$ are adopted to encode instructions and viewpoints respectively. The language encoder encodes the instruction $I$ as a global language feature $X \in \mathbb R^{N_I}$ and the vision encoder encodes the panoramic views and the candidate views as follows:
\begin{equation}
\begin{split}
    X &= \mathrm{E}_{L}(I), \\
    f_t^v = \mathrm{E}_{V}&(V_t), \ \ \ f_t^c = \mathrm{E}_{V}(C_t),
\end{split}
\end{equation}
where $f_t^v$ and $f_t^c$ are features of the current viewpoint and candidates.
A cross-modal attention function $\mathrm{Attn}(\cdot)$~\cite{tan2019lxmert} is introduced to compute visual attention based on textual information. Then a policy network $\pi$ is applied to predict action $a_t$: 
\begin{equation}
\begin{split}
    s_t &= \mathrm{Attn}(X, f_t^v, s_{t-1}), \\
    a_t &= \pi(s_t, f_t^c). 
\end{split}
\end{equation}

\subsection{Contrastive Learning}
In contrastive learning, representations of positive and negative samples are extracted with an encoder $\mathrm{E}(\cdot)$ followed by a mapping function $\mathrm{U}(\cdot)$. For example, $x_i$ is one of $H$ positive examples for the anchor sample $x$. The representation of $x_i$ is denoted by $p_i = \mathrm{U}(\mathrm{E}(x_i))$. For the anchor example $x$, representation is extracted with the encoder $\mathrm{E}(\cdot)$ followed by a projection $\mathrm{U}(\cdot)$ and a predictor $\mathrm{G}(\cdot)$.
Thus, the representation of the anchor $x$ is formulated as $q = \mathrm{G}(\mathrm{U}(\mathrm{E}(x)))$. 
$x_j$ is one of $J$ negative samples, whose representations are $n_{j} = \mathrm{U}(\mathrm{E}(x_j))$. 
Let $\mathcal{P}$ be the set of positive representations and $\mathcal{N}$ be the set of negative representations for each anchor representation $q$. Then for each anchor $q$, we have $\mathcal{P} = \{p_i\}_{i=1}^H$ and $\mathcal{N} = \{n_j\}_{j=1}^J$.
Circle loss~\cite{Sun2020Circle} is one of the embedding losses maximizing within-class similarity and minimizing between-class similarity, and meanwhile updating pair weights more accurately.
It is formulated as:
\begin{equation}
\label{equ:circle}
\mathcal{L}_{\text {circle}}=\log \left[1+\sum_{j=1}^{J} \exp \left(l_n^j\right) \sum_{i=1}^{H} \exp \left(l_p^i\right)\right],
\end{equation}
where logits $l_n^j$ and $l_p^i$ are defined as follows:
\begin{equation}
\begin{split}
l_n^j&=\gamma \left[\mathrm{sim}\left(q, n_j\right)-O_{n}\right]_{+} \left(\mathrm{sim}\left(q, n_j\right)-\Delta_{n}\right), \\
l_p^i&=-\gamma \left[O_{p}-\mathrm{sim}\left(q, p_i\right)\right]_{+} \left(\mathrm{sim}\left(q, p_i\right)-\Delta_{p}\right), 
\label{equ:self_pace}
\end{split}
\end{equation}
where $\gamma$ is a scale factor, $+$ is a cut-off at zero operation, and $\mathrm{sim}$ computes the cosine similarity.
$O_{n}$, $\Delta_{n}$, $O_{p}$ and $\Delta_{p}$ are set as $-m$, $m$, $1+m$ and $1-m$ respectively, where $m$ is the margin for similarity separation. If the similarity score deviates severely from its optimum ($O_{p}$ for positive pairs and $O_{n}$ for negative pairs), it will get a larger weighting factor.

\section{CITL}
In this section, we propose our Contrastive Instruction-trajectory Learning (CITL), consisting of coarse-fine contrastive objectives and a pairwise sample-reweighting mechanism. Fig.~\ref{fig:model} shows the framework of our CITL. 
\subsection{Coarse-grained Contrastive Learning}

Our coarse-grained contrastive learning consists of two contrastive objectives: 1) coarse contrastive loss for trajectories $\mathcal{L}_{C^T}$ and 2) coarse contrastive loss for instructions $\mathcal{L}_{C^I}$. 

\begin{table*}[t]
    \centering
    \caption{Comparison with state-of-the-art methods on R2R. \textbf{Black} indicates best results.}
    \vspace{-1em}
    \resizebox{0.98\textwidth}{!}{
    {\renewcommand{\arraystretch}{1}
    \begin{tabular}{l||cccc|cccc|cccc}
    \specialrule{.1em}{.05em}{.05em}
        \multirow{2}{*}{Methods}
        & \multicolumn{4}{c|}{R2R Val Seen}
        & \multicolumn{4}{c|}{R2R Val Unseen}
        & \multicolumn{4}{c}{R2R Test Unseen} \\
    \cline{2-13}
        & TL & NE$\downarrow$ & SR$\uparrow$ & SPL$\uparrow$ & TL & NE$\downarrow$ & SR$\uparrow$ & SPL$\uparrow$ & TL & NE$\downarrow$ & SR$\uparrow$ & SPL$\uparrow$ \\
    \hline
    \hline
        Random & 9.58 & 9.45 & 16 & - & 9.77 & 9.23 & 16 & - & 9.89 & 9.79 & 13 & 12 \\
        Human & - & - & - & - & - & - & - & - & 11.85 & 1.61 & 86 & 76 \\
    \hline
        Seq2Seq~\cite{Anderson2018Vision} & 11.33 & 6.01 & 39 & - & 8.39 & 7.81 & 22 & - & 8.13 & 7.85 & 20 & 18 \\
        Speaker-Follower~\cite{Fried2018Speaker} & - & 3.36 & 66 & - & - & 6.62 & 35 & - & 14.82 & 6.62 & 35 & 28 \\
        SMNA~\cite{ma2019selfmonitoring} & - & 3.22 & 67 & 58 & - & 5.52 & 45 & 32 & 18.04 & 5.67 & 48 & 35 \\
        RCM+SIL (train)~\cite{Wang2019Reinforced} & 10.65 & 3.53 & 67 & - & 11.46 & 6.09 & 43 & - & 11.97 & 6.12 & 43 & 38 \\
        PRESS~\cite{Li2019RobustNW} & 10.57 & 4.39 & 58 & 55 & 10.36 & 5.28 & 49 & 45 & 10.77 & 5.49 & 49 & 45 \\
        FAST-Short~\cite{Ke2019Tactical} & - & - & - & - & 21.17 & 4.97 & 56 & 43 & 22.08 & 5.14 & 54 & 41 \\
        AuxRN~\cite{Zhu2020Vision} & - & 3.33 & 70 & 67 & - & 5.28 & 55 & 50 & - & 5.15 & 55 & 51 \\
        PREVALENT~\cite{Hao2020Towards} & 10.32 & 3.67 & 69 & 65 & 10.19 & 4.71 & 58 & 53 & 10.51 & 5.30 & 54 & 51 \\
        RelGraph~\cite{Hong2020Language} & 10.13 & 3.47 & 67 & 65 & 9.99 & 4.73 & 57 & 53 & 10.29 & 4.75 & 55 & 52 \\
    \hline
        EnvDrop~\cite{Tan2019Learning} & 11.00 & 3.99 & 62 & 59 & 10.70 & 5.22 & 52 & 48 & 11.66 & 5.23 & 51 & 47 \\
        $+$ CITL & 11.84 & 3.23 & 70 & 66 & 15.47 & 5.06 & 52 & 48 & 10.69 & 5.39 & 54 & 50 \\
        RecBERT (init OSCAR)~\cite{hong2020recurrent} & 10.79 & 3.11 & 71 & 67 & 11.86 &4.29 & 59 & 53 & 12.34 & 4.59 & 57 & 53 \\
        $+$ CITL & 11.22 & 2.99 & 72 & 68 & 15.91 & 4.34 & 60 & 54 & 15.83 & 4.30 & 61 & 55 \\
        RecBERT (init PREVALENT)~\cite{hong2020recurrent} & 11.13 & 2.90 & 72 & 68 & 12.01 & 3.93 & \textbf{63} & 57 & 12.35 & 4.09 & 63 & 57 \\
        $+$ CITL & 11.20 & \textbf{2.65} & \textbf{75} & \textbf{70} & 11.88 & \textbf{3.87} & \textbf{63} & \textbf{58} & 12.30 & \textbf{3.94} & \textbf{64} & \textbf{59} \\
    \specialrule{.1em}{.05em}{.05em}
    \end{tabular}}}
    \label{tab:r2r}
\end{table*}

\begin{table*}[t]
    \centering
    \caption{Comparison with agents on the R4R dataset. $^*$ indicates results we reproduce. \textbf{Black} indicates best results.}
    \vspace{-1 em}
    \resizebox{0.98\textwidth}{!}{
    {\renewcommand{\arraystretch}{1}
    \begin{tabular}{l||cccccc|cccccc}
    \specialrule{.1em}{.05em}{.05em}
        \multirow{2}{*}{Methods}
        & \multicolumn{6}{c|}{R4R Val Seen}
        & \multicolumn{6}{c}{R4R Val Unseen} \\
    \cline{2-13}
        & NE$\downarrow$ & SR$\uparrow$ & SPL$\uparrow$ & CLS$\uparrow$ & nDTW$\uparrow$ & SDTW$\uparrow$ & NE$\downarrow$ & SR$\uparrow$ & SPL$\uparrow$ & CLS$\uparrow$ & nDTW$\uparrow$ & SDTW$\uparrow$ \\
    \hline
    \hline
        Speaker-Follower~\cite{Fried2018Speaker} & 5.35 & 51.9 & 37.3 & 46.4 & - & - & 8.47 & 23.8 & 12.2 & 29.6 & - & - \\
        RCM (goal)~\cite{Wang2019Reinforced} & 5.11 & 55.5 & 32.3 & 40.4 & - & - & 8.45 & 28.6 & 10.2 & 20.4 & - & - \\
        RCM (fidelity)~\cite{Wang2019Reinforced} & 5.37 & 52.6 & 30.6 & 55.3 & - & - & 8.08 & 26.1 & 7.7 & 34.6 & - & - \\
        PTA high-level~\cite{Federico2019Perceive} & 4.54 & 58 & 39 & \textbf{60} & \textbf{58} & 41 & 8.25 & 24 & 10 & 37 & 32 & 10 \\
        EGP~\cite{Deng2020Evolving} & - & - & - & - & - & - & 8.00 & 30.2 & - & 44.4 & 37.4 & 17.5 \\
        BabyWalk~\cite{zhu2020babywalk} & - & - & - & - & - & - & 8.2 & 27.3 & 14.7 & \textbf{49.4} & \textbf{39.6} & 17.3 \\
    \hline
        RecBERT (init PREVALENT)$^*$ \cite{hong2020recurrent} & 4.27 & 60.5 & 51.9 & 53.3 & 51.6 & 37.7 & 6.73 & 41.2 & 31.7 & 39.6 & 36.8 & 21.6 \\
        $+$ CITL & \textbf{3.48} & \textbf{66.8} & \textbf{57.0} & 56.4 & 55.2 & \textbf{42.7} & \textbf{6.42} & \textbf{44.4} & \textbf{35.1} & 39.6 & 37.4 & \textbf{23.4} \\
    \specialrule{.1em}{.05em}{.05em}
    \end{tabular}}}
    \label{tab:r4r}
\end{table*}

\begin{figure}
    \centering
    \includegraphics[width=0.85\linewidth]{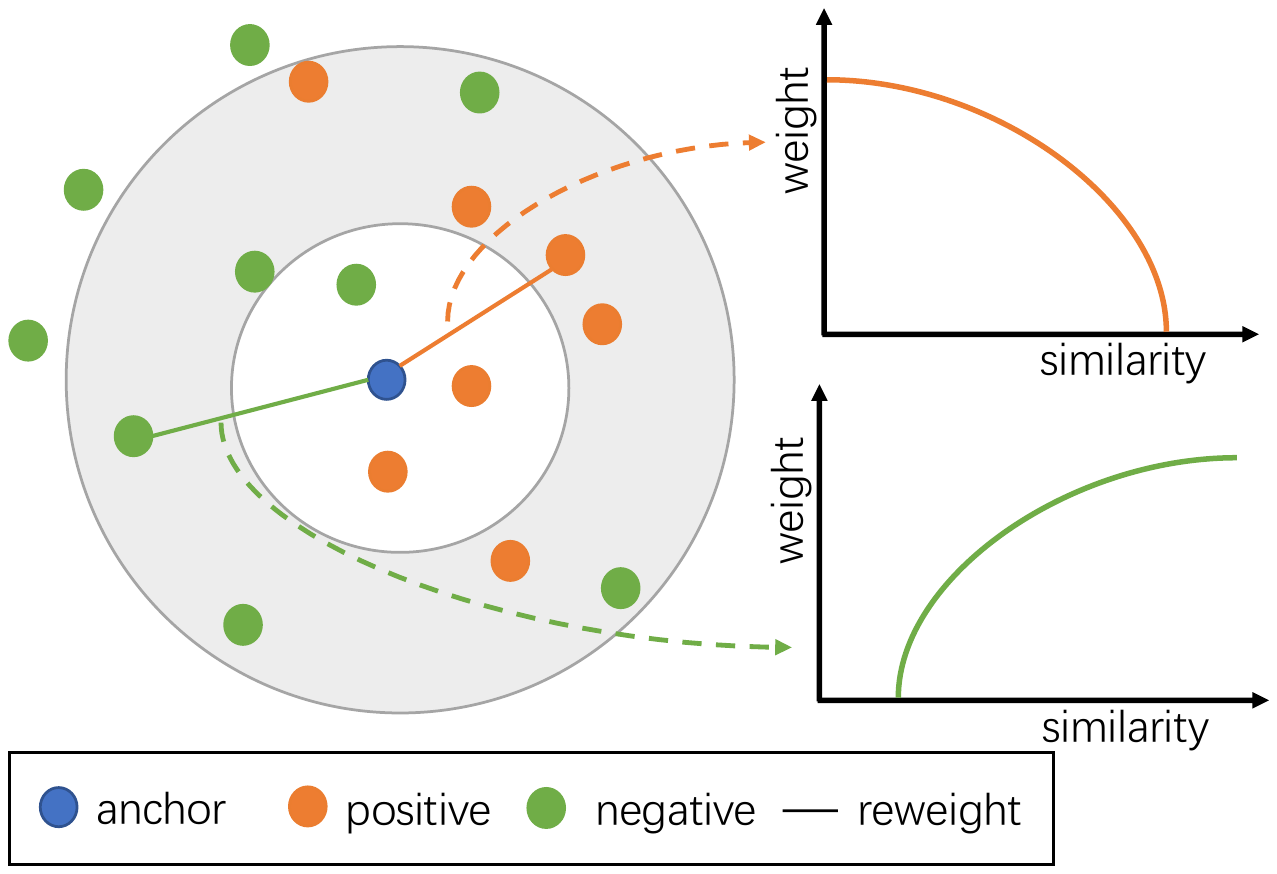}
    \caption{Illustration of the pairwise sample-reweighting mechanism. Samples distributed in white areas are easy samples or false-negative samples. Correlations of similarities and weights in the loss are shown at the right (top for positive pairs and bottom for negative pairs).}
    \vspace{-1 em}
    \label{fig:sample_reweight}
\end{figure}

\noindent
\textbf{Trajectory Loss}
The optimal trajectory is the shortest path from the starting position to the ending position.  
Learning from only the optimal trajectories may lead to over-fitting problems since the optimal trajectories only occupy a small proportion of the feasible navigation trajectories. To alleviate the over-fitting problems, we propose to learn not only from optimal trajectories but also from sub-optimal trajectories.
As shown in Fig.~\ref{fig:model}, we define sub-optimal trajectories as the ones that have the same starting and ending points as the optimal trajectory and their lengths are shorter than a threshold.
Positive sub-optimal trajectories should be close to the anchor, while intra-negative ones should deviate heavily from the anchor.
The hop (step count) of a sub-optimal trajectory $T$ is denoted as $h(T)$, and the hop of the optimal one is denoted as $h_{gt}$.
We introduce two hyper-parameters $\alpha_p$ and $\alpha_n$ ($2 \textgreater \alpha_n \textgreater \alpha_p \textgreater 1$) to separate these trajectories into positive samples and intra-negative samples: 
\begin{equation}
\begin{split}
 \mathcal{P}^T = \{T_i | h(T_i) \leq \alpha_p \cdot h_{gt} \}, \\ 
 \mathcal{N}^T = \{T_i | h(T_i) \geq \alpha_n \cdot h_{gt} \},
 \end{split}
\end{equation}
In this way, we get all initial positive trajectories $\mathcal{P}^T$ and intra-negative trajectories $\mathcal{N}^T$.
To help the model distinguish different instances and improve efficiency, we introduce a memory bank $\mathcal{M}^T$ to make use of representations of inter-negative samples from previous batches. Then in the current batch, we get all negative representations by unifying intra-negative trajectories and inter-negative samples:
\begin{equation}
\mathcal{N}_{full}^T = \mathcal{N}^T \cup \mathcal{M}^T. 
\label{equ:full_neg}
\end{equation}

Therefore the coarse contrastive loss for trajectories is formulated as:
\begin{equation}
\mathcal{L}_{C^T} = \mathcal{L}_{\text {circle}}\left( q, \mathcal{P}^T, \mathcal{N}_{full}^T \right).
\end{equation}
After computing $\mathcal{L}_{C^T}$, the memory bank $\mathcal{M}^T$ is updated by replacing oldest positive samples with $\mathcal{P}^T$.

\noindent
\textbf{Instruction Loss}
Natural languages contain considerable noise due to their diversity, like multiple synonyms.
To overcome this problem, we implement a contrastive objective for instruction-level comparison among the diversified language descriptions.
First of all, we adopt three natural language processing augmentation methods to generate high-quality positive instructions given a query instruction: 1) using the WordNet to substitute words with their \textbf{synonyms}~\cite{Zhang2015Character}; 2) using a pre-trained BERT to \textbf{insert} or \textbf{substitute} words according to context~\cite{Anaby2020Do,kumar2020Data}; and 3) \textbf{back-translation}~\cite{Xie2020Unsupervised}. We assume that the augmented instructions should preserve semantic information of the original ones. 
To obtain the intra-negative instruction of the query instruction, we first generate sub-instructions as in~\cite{hong2020sub}. 
These sub-instructions are shuffled or repeated randomly and then reassembled to become an intra-negative instruction.
All augmented samples are fed into the language encoder $\mathrm{E}_{L}$ to get positive and intra-negative language representations. 
After that, we get the positive representations $\mathcal{P}^I$ and the intra-negative representations $\mathcal{N}^I$. We also introduce a memory bank $\mathcal{M}^I$ for instruction to store inter-negative representations. Then we unify $\mathcal{N}^I$ and $\mathcal{M}^I$ to get full negative set $\mathcal{N}_{full}^I$ following Eq.~\ref{equ:full_neg}.
For each positive instruction representation $q$, the coarse contrastive loss for instruction is defined as:
\begin{equation}
\mathcal{L}_{C^I} = \mathcal{L}_{\text {circle}}\left( q, \mathcal{P}^I, \mathcal{N}_{full}^I \right).
\end{equation}
Similar to $\mathcal{L}_{C^T}$, the memory bank $\mathcal{M}^I$ is updated with $\mathcal{P}^I$.

\begin{table}[t]
    \centering
    \caption{Comparison on the RxR monolingual unseen validation set. $^*$ indicates results we reproduce (init PREVALENT). \textbf{Black} indicates best results.}
    \vspace{-1 em}
    \resizebox{1.0\linewidth}{!}{
    \begin{tabular}{l|ccccc}
    \specialrule{.1em}{.05em}{.05em}
        Model & SR$\uparrow$ & SPL$\uparrow$ & CLS$\uparrow$ & nDTW$\uparrow$ & SDTW$\uparrow$ \\
    \hline
        EnvDrop~\cite{Tan2019Learning} & 38.5 & 34 & 54 & 51 & 32 \\
        Syntax~\cite{li2021improving} & 39.2 & 35 & 56 & 52 & 32 \\
    \hline
        RecBERT$^*$~\cite{hong2020recurrent} & 44.9 & 39.3 & 56.2 & 52.5 & 36.3 \\
        $+$ CITL & \textbf{47.2} & \textbf{40.7} & \textbf{56.9} & \textbf{53.5} & \textbf{37.6} \\
    \specialrule{.1em}{.05em}{.05em}
    \end{tabular}}
    \label{tab:rxr}
\end{table}

\begin{table*}[t]
    \centering
    \caption{Results on semi-supervised evaluation (trained with 1\%, 5\% and 10\% training data). $^*$ indicates results we reproduce.}
    \vspace{-1 em}
    \resizebox{1.0\textwidth}{!}{
    {\renewcommand{\arraystretch}{1}
    \begin{tabular}{l||cccc|cccc|cccc}
    \specialrule{.1em}{.05em}{.05em}
        \multirow{2}{*}{Model}
        & \multicolumn{4}{c|}{R2R Val Unseen (1\%)}
        & \multicolumn{4}{c|}{R2R Val Unseen (5\%)}
        & \multicolumn{4}{c}{R2R Val Unseen (10\%)} \\
    \cline{2-13}
        & TL & NE$\downarrow$ & SR$\uparrow$ & SPL$\uparrow$ & TL & NE$\downarrow$ & SR$\uparrow$ & SPL$\uparrow$ & TL & NE$\downarrow$ & SR$\uparrow$ & SPL$\uparrow$ \\
    \hline
    \hline
        RecBERT (init OSCAR)$^*$~\cite{hong2020recurrent} & 8.69 & 9.08 & 17.79 & 16.49 & 9.77 & \textbf{8.24} & 24.39 & 22.48 & 10.74 & 7.46 & 31.89 & 29.14 \\
        $+$ CITL & 10.03 & \textbf{8.90} & \textbf{18.90} & \textbf{17.31} & 10.31 & 8.35 & \textbf{25.67} & \textbf{23.34} & 10.21 & \textbf{7.10} & \textbf{34.48} & \textbf{31.47} \\
    \hline
        RecBERT (init PREVALENT)$^*$~\cite{hong2020recurrent} & 13.77 & 7.49 & \textbf{32.65} & 27.96 & 11.56 & 6.07 & 42.32 & 37.67 & 12.86 & 5.37 & 48.11 & 42.69 \\
        $+$ CITL & 11.13 & \textbf{7.18} & 32.52 & \textbf{29.00} & 11.23 & \textbf{5.72} & \textbf{45.38} & \textbf{41.54} & 12.10 & \textbf{5.28} & \textbf{50.02} & \textbf{45.05} \\
    \specialrule{.1em}{.05em}{.05em}
    \end{tabular}}}
    \vspace{-1 em}
    \label{tab:semisupervise}
\end{table*}

\subsection{Fine-grained Contrastive Learning}

The coarse-grained contrastive learning focuses on whole trajectories and instructions. In contrast, the fine-grained contrastive loss focuses on sub-instructions and introduces temporal information to help the agent analyze the coherence of sub-instructions. Here we propose a fine-grained contrastive loss for sub-instructions $\mathcal{L}_{F^I}$.

\noindent\textbf{Sub-instruction Loss}
We propose a fine-grained contrastive strategy for sub-instructions to help the agent learn the temporal information of sub-instructions and analyze instructions better. We assume that adjoining sub-instructions have a sense of coherence. Thus their representations should be similar to some degree. Those sub-instructions which are not neighbors should be pulled apart. To generate positive and intra-negative samples, we first generate sub-instructions given instruction as in~\cite{hong2020sub}. Then we randomly select a sub-instruction as the query sub-instruction.
The nearest neighbors of this query sub-instruction are positive samples, while others are intra-negative samples.
Similar to the coarse contrastive losses, a memory bank $\mathcal{M}^{SI}$ is introduced to store inter-negative sub-instructions from other instructions. Similar to the instruction loss $\mathcal{L}_{C^I}$, positive and intra-negative language representations are extracted via the language encoder $\mathrm{E}_{L}$.
For the query sub-instruction $q$, the fine-grained contrastive loss is formulated as:
\begin{equation}
\mathcal{L}_{F^I} = \mathcal{L}_{\text {circle}}\left( q, \mathcal{P}^{SI}, \mathcal{N}_{full}^{SI} \right).
\end{equation}
The memory bank $\mathcal{M}^{SI}$ is updated with $\mathcal{P}^{SI}$.

\subsection{Pairwise Sample-reweighting Mechanism}
There are large amounts of easy samples in augmented samples and memory banks, causing the training to plateau quickly and occupy extensive memory usage. 
To alleviate this, we propose the pairwise sample-reweighting mechanism equipped with a novel pair mining strategy to explore hard samples and reweight different pairs. The overview is shown in Fig.~\ref{fig:sample_reweight}.

\noindent\textbf{Pair Mining}
We introduce our pair mining strategy that aims to select informative samples and discard less informative ones. For the anchor $q$, positive and negative sets are denoted as $P$ and $N$. Negative samples are selected as follows compared with the hardest positive sample:
\begin{small}
\begin{equation}
S_n = \{n_j | 1 - m > \mathrm{sim}\left(q, n_j\right) > \min \{\mathrm{sim}\left( q, p_i \right) \} - m \},
\label{equ:mine1}
\end{equation}
\end{small}
where $p_i \in \mathcal{P}$ and $n_j \in \mathcal{N}$. If the similarity score is greater than $1-m$, this negative sample will be regarded as false negative and then discarded. After selecting negative samples, positive samples are compared with the remaining hardest negative sample:
\begin{equation}
S_p = \{p_i | \mathrm{sim}\left(q, p_i\right) < \max \{\mathrm{sim}\left(q, n_j\right) \} + m \},
\label{equ:mine2}
\end{equation}
where $n_j \in S_n(\mathcal{N})$.

\noindent\textbf{Sample reweighting}
The remaining samples will be reweighted by self-paced reweighting following Eq.~\ref{equ:self_pace}. Unlike previous methods, our sample reweighting focuses on sequantial data. The reweighted loss can be formulated as $\mathcal{L}_{\text {circle}} (q, S_p(\mathcal{P}), S_n(\mathcal{N}))$.
Hence the coarse-fine contrastive losses are rewrited as follows:
\begin{equation}
\begin{split}
\mathcal{L}_{C^T}^{'} &= \mathcal{L}_{C^T} (q, S_p(\mathcal{P}^T), S_n(\mathcal{N}_{full}^T)), \\
\mathcal{L}_{C^I}^{'} &= \mathcal{L}_{C^I} (q, S_p(\mathcal{P}^I), S_n(\mathcal{N}_{full}^I)), \\
\mathcal{L}_{F^I}^{'} &= \mathcal{L}_{F^I} (q, S_p(\mathcal{P}^{SI}), S_n(\mathcal{N}_{full}^{SI})).
\end{split}
\end{equation}

\subsection{Training}
We train the model with a mixture of contrastive learning, reinforcement learning (RL) and imitation learning (IL). The agent learns by following teacher actions $a_t^*$:
\begin{equation}
\mathcal{L}_{IL} = \sum_t -a_t^*log\left(p_t\right).
\end{equation}
RL is adopted to avoid overfitting in VLN. Here we adopt A2C~\cite{Volodymyr2016Asynchronous} algorithm. The loss function is formulated as: 
\begin{equation}
\mathcal{L}_{RL} = -\sum_t a_t log\left(p_t\right) A_t.
\end{equation}
$p_t$ and $A_t$ are predicted logits and the advantage function. The full loss in our proposed model is as:
\begin{equation}
\mathcal{L} = \mathcal{L}_{IL} + \mathcal{L}_{RL} + \lambda_1 \mathcal{L}_{C^P}^{'} + \lambda_2 \mathcal{L}_{C^I}^{'} + \lambda_3 \mathcal{L}_{F^I}^{'}.
\end{equation}
where $\lambda_1$, $\lambda_2$ and $\lambda_3$ are weighting factors.

\section{Experiments}

\noindent\textbf{Datasets}
We evaluate the CITL on several popular VLN datasets.
The R2R~\cite{Anderson2018Vision} dataset consists of 90 housing environments. The training set comprises 61 scenes, and the validation unseen set and test unseen set contain 11 and 18 scenes respectively.
R4R~\cite{Vihan2019Stay} concatenates the trajectories and instructions in R2R. 
RxR~\cite{ku2020room} is a larger dataset containing more extended instructions and trajectories.

\noindent\textbf{Experimental Setup}
All experiments are conducted on an NVIDIA 3090 GPU. We also use the MindSpore Lite tool~\cite{mindspore}. In all contrastive losses, the margin $m$ is set to 0.25, and $\lambda_1$, $\lambda_2$ and $\lambda_3$ are fixed to 0.1, 0.01 and 0.01 respectively. The size of all memory banks is fixed to 240. $\alpha_p$ and $\alpha_n$ are set to 1.2 and 1.4 respectively. Training schedules are the same as baselines~\cite{Tan2019Learning,hong2020recurrent}. %
We use the same augmentation data as in~\cite{Hao2020Towards} when adopting RecBERT~\cite{hong2020recurrent} as the baseline.

\noindent\textbf{Evaluation Metrics}
For R2R, the agent is evaluated using the following metrics~\cite{Peter2018Evaluation,Anderson2018Vision}: Trajectory Length (TL), Navigation Error (NE), Success Rate (SR) and Success weighted by Path Length (SPL).  Additional metrics are used for R4R and RxR, including Coverage weighted by Length Score (CLS)~\cite{Vihan2019Stay} and Normalized Dynamic Time Warping (nDTW)~\cite{Gabriel2019General} and Success rate weighted normalized Dynamic Time Warping (SDTW)~\cite{Gabriel2019General}.

\subsection{Comparison with SoTA}
Results in Table~\ref{tab:r2r} compare the single-run (greedy search, no pre-exploration~\cite{Wang2019Reinforced}) performance of different agents on the R2R benchmark. Our base model initialised from PREVALENT~\cite{Hao2020Towards}, a pre-trained model for VLN, performs better than previous methods~\cite{hong2020recurrent} over all dataset splits, achieving 59\% SPL ($+$2\%) on the test set.
Comparing to previous methods, we can see that the improvement of the test set is greater than the unseen validation split, which suggests the strong generalization of our agent by equipping with coarse/fine-grained semantic contrast. Table~\ref{tab:r4r} shows results on the R4R dataset. Our CITL performs consistently better than the RecBERT baseline, showing that our model can generalize well to long instruction and trajectory. 
Table~\ref{tab:rxr} compares CITL with previous state-of-the-art on the RxR dataset. 
Our model gets significant improvement ($+$1.4\% in SPL and $+$2.3\% in SR) compared with its RecBERT backbone, and outperforms previous state-of-the-art models on all metrics.  

\begin{figure}
    \centering
    \includegraphics[width=0.8\linewidth]{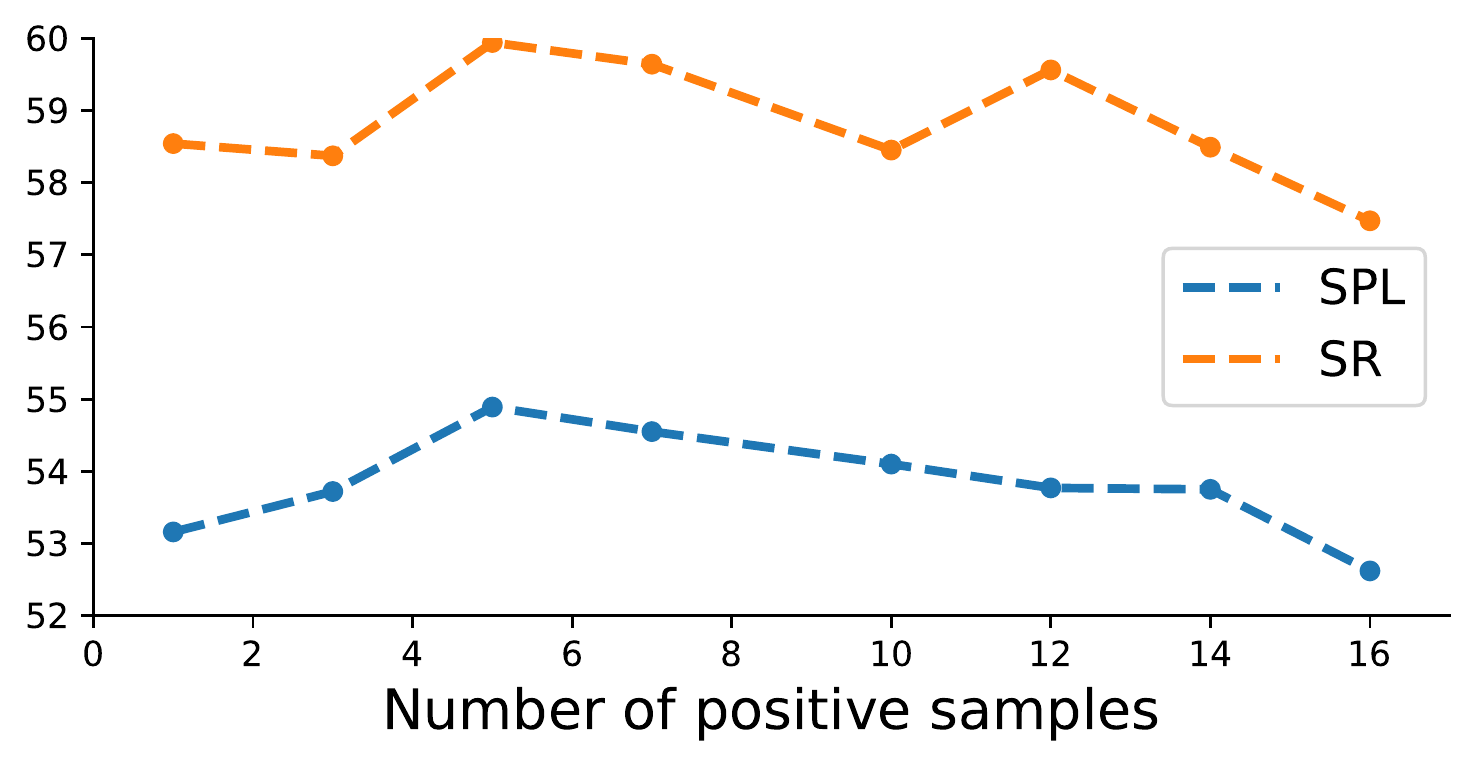}
    \caption{Ablation study on multi-pair NCE loss $\mathcal{L}_{\mathrm{mul}}$ with the coarse contrastive strategy for trajectories.}
    \vspace{-1 em}
    \label{fig:nce_pos}
\end{figure}

\subsection{Alation Study}
We further study the effectiveness of each component of CITL over the R2R dataset without augmentation data generated by the speaker.

\noindent\textbf{Semi-Supervised Evaluation}
To validate the robustness of the proposed method and the ability to acquire exceptional knowledge with less training data, we conduct some experiments in the semi-supervised setting, in which we train the agent with only 1\%, 5\% and 10\% of the training data. 
Table~\ref{tab:semisupervise} presents results on the validation unseen split. 
Our CITL achieves better on SPL under all semi-supervised settings. Notably, the agent initialized from PREVALENT improves consistently over the baseline (1.04\%, 3.87\% and 2.36\% absolute improvements with 1\%, 5\% and 10\% training data respectively).

\noindent\textbf{Common Contrastive Loss}
We first investigate the common InfoNCE loss~\cite{oord2018representation}, which does not reweight samples explicitly. As shown in Fig.~\ref{fig:nce_pos}, choosing multi-pair InfoNCE loss to implement the trajectory loss causes the result susceptible to the number of augmented trajectories. For example, the model performs best with 5 sub-optimal trajectories, but it suffers when the number is increasing or decreasing.

\noindent\textbf{Pairwise Sample-reweighting  Mechanism}
We present detailed comparisons on each module to validate our pairwise sample-reweighting mechanism as in Table~\ref{tab:loss}.
Our proposed pairwise sample-reweighting mechanism performs better than multi-pair InfoNCE loss (55.47\% vs. 52.62\% SPL). Simply using circle loss~\cite{Sun2020Circle} as the contrastive loss does not help the agent fully leverage semantic information. Adding a memory bank can store more samples for contrastive learning, but many easy samples and some noisy data harm the training. Thus, pair mining to select hard samples improves the agent's performance (53.14\% to 55.47\% SPL). This evidence confirms that hard positive and negative samples are crucial in our contrastive losses. 

We also conduct experiments on InfoNCE loss with pair mining in Table~\ref{tab:loss}. The number of positive samples is set to 16 to get better results in pair mining. We can see that it can improve the performance of InfoNCE loss. However, the final result is worse than our pairwise sample-reweighting mechanism since InfoNCE loss cannot reweight hard and easy samples differently.

\begin{table}[t]
    \centering
    \caption{Ablation study on NCE/circle loss and pairwise sample-reweighting mechanism in trajectory loss $\mathcal{L}_{C^T}$. $\mathcal{M}$ is the memory bank for trajectories, and PM is pair mining.}
    \vspace{-1 em}
    \resizebox{1.0\linewidth}{!}{
    {\renewcommand{\arraystretch}{1}
    \begin{tabular}{l||cccc|cccc}
    \specialrule{.1em}{.05em}{.05em}
        \multirow{2}{*}{Loss}
        & \multicolumn{4}{c|}{Module}
        & \multicolumn{4}{c}{R2R Val Unseen} \\
    \cline{2-9}
        & $\mathcal{L}_{\mathrm{mul}}$ & $\mathcal{L}_{\text {circle}}$ & $\mathcal{M}$ & PM & TL & NE$\downarrow$ & SR$\uparrow$ & SPL$\uparrow$ \\
    \hline
    \hline
        \ding{172} & \checkmark & & \checkmark & & 11.32 & 4.45 & 57.47 & 52.62 \\ 
        \ding{173} & \checkmark & & \checkmark & \checkmark & 11.74 & 4.30 & 59.17 & 54.10 \\
    \hline
        \ding{174} &  & \checkmark & & & 12.12 & 4.28 & 59.05 & 53.35 \\ 
        \ding{175} &  & \checkmark & \checkmark & & 11.37 & 4.44 & 58.49 & 53.14 \\ 
        \ding{176} &  & \checkmark &  & \checkmark & 11.92 & \textbf{4.11} & 59.98 & 54.54 \\ 
    \hline
        Full &  & \checkmark & \checkmark & \checkmark & 11.70 & 4.29 & \textbf{60.90} & \textbf{55.47} \\
    \specialrule{.1em}{.05em}{.05em}
    \end{tabular}}}
    \label{tab:loss}
\end{table}

\begin{table}[t]
    \centering
    \caption{Results on different coarse-fine contrastive losses.}
    \vspace{-1 em}
    \resizebox{1.0\linewidth}{!}{
    {\renewcommand{\arraystretch}{1}
    \begin{tabular}{l||ccc|cccc}
    \specialrule{.1em}{.05em}{.05em}
         \multirow{2}{*}{Models}
         & \multicolumn{3}{c|}{Losses}
         & \multicolumn{4}{c}{R2R Val Unseen} \\
    \cline{2-8}
         & $\mathcal{L}_{C^T}^{'}$ & $\mathcal{L}_{C^I}^{'}$ & $\mathcal{L}_{F^I}^{'}$ & TL & NE$\downarrow$ & SR$\uparrow$ & SPL$\uparrow$ \\
    \hline
    \hline
        Baseline & & & & 10.99 & 4.47 & 57.17 & 52.90 \\
    \hline
        \ding{172} & \checkmark & & & 11.70 & 4.29 & 60.90 & 55.47 \\
        \ding{173} & & \checkmark & & 12.23 & 4.22 & 61.00 & 55.17 \\
        \ding{174} & & & \checkmark & 11.63 & 4.37 & 58.24 & 53.58 \\
    \hline
        Full & \checkmark & \checkmark & \checkmark & 12.36 & \textbf{3.98} & \textbf{62.11} & \textbf{55.83} \\
    \specialrule{.1em}{.05em}{.05em}
    \end{tabular}}}
    \label{tab:strategy}
\end{table}

\noindent\textbf{Coarse/fine-grained Contrastive Losses}
Table~\ref{tab:strategy} shows comprehensive ablation experiments on our coarse/fine-grained contrastive losses. As the results suggested, employing coarse contrastive losses leads to substantial performance gains, which suggests that exploiting the semantics of the cross-instance instruction-trajectory pairs in contrastive learning improves navigation. Meanwhile, employing fine-grained contrastive loss to learn temporal information of sub-instructions also enhances the performance, which indicates that the agent may benefit from analyzing relations of sub-instructions. Combining coarse/fine-grained contrastive loss further improves the agent's performance (52.90\% to 55.83\% SPL).

\section{Conclusion}
In this paper, we propose a novel framework named CITL, with coarse/fine-grained contrastive learning. Coarse-grained contrastive learning fully explores the semantics of cross-instance samples and enhances vision-and-language representations to improve the performance of the agent. The fine-grained contrastive learning learns to leverage the temporal information of sub-instructions. The pairwise sample-reweighting mechanism mines hard samples and eliminates the effects of false-negative samples, hence mitigating the influence of augmentation bias and improving the robustness of the agent. Our CITL achieves promising results, which indicates the robustness of the model.

\section*{Acknowledgement}
This work was supported in part by National Key R\&D Program of China under Grant No. 2020AAA0109700, National Natural Science Foundation of China (NSFC) under Grant No.U19A2073 and No.61976233, Guangdong Province Basic and Applied Basic Research (Regional Joint Fund-Key) Grant No.2019B1515120039, Guangdong Outstanding Youth Fund (Grant No. 2021B1515020061), Shenzhen Fundamental Research Program (Project No. RCYX20200714114642083, No. JCYJ20190807154211365) and CAAI-Huawei MindSpore Open Fund. We thank MindSpore for the partial support of this work, which is a new deep learning computing framwork\footnote{https://www.mindspore.cn/}.


\end{document}